\documentclass[a4paper,conference]{IEEEtran}

\usepackage{cite}

\ifCLASSINFOpdf
\else
\fi

\usepackage{url}

\usepackage{times}
\usepackage{epsfig}
\usepackage{graphicx}
\usepackage{amsmath}
\usepackage{amssymb}
\usepackage{subfigure}

\usepackage{mathtools}
\DeclareMathOperator*{\argmin}{argmin}

\usepackage[linesnumbered,ruled]{algorithm2e}

\begin{document}

\title{Deep Structured Energy-Based Image Inpainting}

\author{\IEEEauthorblockN{Fazil Altinel\IEEEauthorrefmark{1}, Mete Ozay\IEEEauthorrefmark{1}, and Takayuki Okatani\IEEEauthorrefmark{1}\IEEEauthorrefmark{2}} \IEEEauthorblockA{\IEEEauthorrefmark{1}Graduate School of Information Sciences, Tohoku University, Sendai,  Japan\\
\IEEEauthorblockA{\IEEEauthorrefmark{2}RIKEN Center for AIP, Tokyo, Japan}
Email: \{altinel, mozay, okatani\}@vision.is.tohoku.ac.jp}
}

\maketitle

\begin{abstract}
In this paper, we propose a structured image inpainting method employing an energy based model. In order to learn structural relationship between patterns observed in images and missing regions of the images, we employ an energy-based structured prediction method. The structural relationship is learned by minimizing an energy function which is defined by a simple convolutional neural network. The experimental results on various benchmark datasets show that our proposed method significantly outperforms the state-of-the-art methods which use Generative Adversarial Networks (GANs). We obtained 497.35 mean squared error (MSE)  on the Olivetti face dataset compared to 833.0 MSE provided by the state-of-the-art method. Moreover, we obtained 28.4 dB peak signal to noise ratio (PSNR) on the SVHN dataset and 23.53 dB on the CelebA dataset, compared to 22.3 dB and 21.3 dB, provided by the state-of-the-art methods, respectively. The code is publicly available.\footnote{ \url{https://github.com/cvlab-tohoku/DSEBImageInpainting}.}
\end{abstract}


%
\IEEEpeerreviewmaketitle

\section{Introduction}
\label{sec:introduction}

In this work, we address an image inpainting problem, which is to recover missing regions of an image, as shown in Fig.1. The key to the problem is how to model structural relationship between local regions of natural images, and then use it for estimating pixel intensities of the missing regions. Indeed, this has been a major concern in previous studies of image inpainting, and therefore it is often formulated as a structured prediction problem~\cite{HighResolutionCVPR, LearningDeepStructured}. In general, structured prediction problems are formulated as energy minimization; an energy function is incorporated to model structural relationship between unknowns (e.g., pixel intensities in the case of image inpainting), and the unknowns are estimated by minimizing it.

Recently, deep networks were applied to image inpainting, and have achieved great success~\cite{pathakCVPR16context, yeh2017semantic, GAN}. In these works, deep networks are trained using training samples so that they can estimate pixel intensities of  missing regions of an input image; the computation is performed in a feedforward manner from an input image to an output inpainted image. In the training step, generative adversarial networks (GANs) are popularly used. In GANs which do not employ  energy functions, the structural relationship between image pixels is implicitly represented internally inside the deep networks, which is obtained through learning (and partly modeled by the network architectures). Likewise, similar deep-network-based approaches have been applied with success to various problems which were formerly treated as structured prediction problems, such as human pose estimation, image super-resolution, etc.

In parallel to this trend, Belanger et al.~\cite{BelangerSPEN} proposed to use deep networks in the framework of structured prediction, where deep networks are used to define energy functions. As compared with previous approaches such as manually designed energy functions or training of shallow models (e.g., conditional random fields (CRFs)), their approach is expected to improve the representation power of energy functions as well as their accuracy in modeling structural relationship through learning from data. The structural relationship thus learned is still implicitly represented internally in deep networks. However, the performance of this method to various structured prediction problems is not so clear as of now; for instance, it remains unclear which approach is better for individual problems between the energy-based and the non-energy-based methods, both of which use deep networks. This must and can only be clarified through evaluations by its applications to various tasks.

\begin{figure}[t]
\begin{center}
\includegraphics[width=0.9\linewidth]{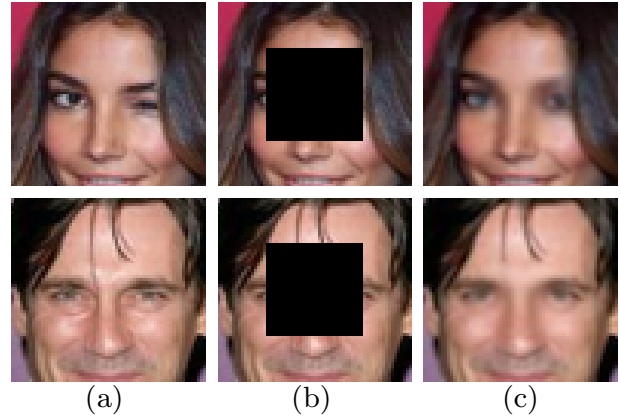}
\end{center}
  \caption{An illustration of image inpainting results on face images. Given images with missing region (b), our method generates visually realistic and consistent inpainted images (c). (a) Ground truth images, (b) occluded input images, (c) inpainting results by our method.}
\label{fig:introduction}
\end{figure}

With these in mind, in this study, we apply the energy-based method that uses deep networks to image inpainting, while there is no study in the literature to the authors’ knowledge. As shown in Fig.~\ref{fig:introduction}, we aim to learn the structural relationship between the unmasked and masked regions in a natural image. For this purpose, we propose a two-path CNN having inter-layer connections between the paths, which takes an input image with missing regions and any image (which is to be optimized in the inference phase) as the inputs of the two paths, respectively. It outputs energy for the input image pair from its final layer. In the training phase, we train this CNN using a dataset of a pair of an original image and its masked version. The trained CNN provides an energy function that models \textit{deep structural relationship} between unmasked and masked regions for different images. In the inference phase, given an image with masked regions, we minimize the energy function realized by the CNN, yielding an inpainted image in which masked regions are filled with estimated pixels (Section~\ref{sec:method}).

We tested the proposed method on benchmark image inpainting datasets~\cite{celeba, Olivetti, svhn}. We evaluate the proposed method qualitatively and quantitatively, and show its success compared to the state-of-the-art methods~\cite{pathakCVPR16context, yeh2017semantic} in Section~\ref{sec:results}. The results show that our proposed method successfully generates both realistic and accurate inpainted images for given masked images. Our contributions are summarized as follows:
\begin{itemize}
    \item We propose a simple yet efficient deep structured image inpainting method. In order to learn the structural relationship between image pixels in the proposed method, we suggest an energy based model which employs an energy function defined by a CNN.
    \item We demonstrate results of our method on various benchmark image inpainting datasets to show the effectiveness of our method. Our proposed method obtains state-of-the-art performance, and generates more realistic and accurate inpainted images compared to the baseline methods.
\end{itemize}

\section{Related Work}
\label{sec:rw}
Recently, deep learning has been employed for image inpainting and provided remarkable results~\cite{liu2016learning, HighResolutionCVPR, onDemand}. The methods which use Generative Adversarial Networks (GANs)~\cite{GAN} achieved promising results on benchmark image inpainting tasks~\cite{pathakCVPR16context, IizukaSIGGRAPH2017, yeh2017semantic}. 

Pathak et al.~\cite{pathakCVPR16context} proposed an encoder-decoder network for image inpainting called Context Encoders (CEs). CEs are trained by minimizing a function of $ \ell_2 $ loss and adversarial loss~\cite{GAN} to inpaint occluded image regions. Although CEs are able to generate promising inpainting results, they make use of structures of occluded regions during training but not during inference. Thus, inpainting results of CEs are sometimes visually unrealistic. 

Yeh et al.~\cite{yeh2017semantic} proposed a semantic image inpainting framework which includes contextual and perceptual losses. Their framework leverages a pre-trained Deep Convolutional GAN (DCGAN)~\cite{dcgan}, and aims to find the closest mapping in the latent space. However, inpainting results of their framework usually have differences in color along mask boundaries. Therefore, some post-processing methods, such as Poisson blending~\cite{poisssonBlending}, are used to eliminate color differences after inpainted images are generated by the framework. Moreover, their method requires many iterations to find the closest mapping in the latent space during testing phase. On the other hand, our method can generate realistic and visually consistent contents, and requires no post-processing. 

Iizuka et al.~\cite{IizukaSIGGRAPH2017} proposed an approach for image completion using deep neural networks (DNNs). Their method is {\em non-blind inpainting}, that is, it requires not only an input image but a mask indicating regions to be inpainted. The above two methods~\cite{yeh2017semantic, pathakCVPR16context} including ours perform blind inpainting where no mask is necessary. 

In addition, DNNs have been used for structured prediction due to its generative power~\cite{LearningDeepStructured, TrainingDeepNeuralNets, GAN, DRAW, GRAN, HighResolutionCVPR, GenerativeLSTM}. Predicting structured outputs using energy based learning was analyzed in~\cite{Lecun06atutorial}. Belanger et al.~\cite{BelangerSPEN} proposed a structured prediction energy network (SPEN) in which a DNN is exploited to define an energy function, and predictions are obtained by minimizing the energy function using gradient descent. Belanger et al. \cite{BelangerEndtoEndSPEN} introduced an end-to-end learning method for SPENs, where an energy function is minimized by back-propagation using gradient-based prediction. 

Amos et al.~\cite{ICNN} proposed input convex neural networks (ICNNs) which share similar architectural properties compared to SPENs. ICNNs add constraints to the parameters of  SPENs, such that their energy function is convex with respect to some parameters of the energy function and the optimization can be performed globally. Our architecture is similar to that of SPENs and ICNNs. Unlike SPENs, which only consider multi-label classification problems, we address image inpainting problems. Employment of end-to-end learning methods enables SPENs to handle more complex tasks such as depth image denoising. 

We employ a specific CNN which includes connections from input layer to hidden layers in order to achieve realistic image inpainting results. Such connections have been recently discussed in deep residual networks~\cite{deepResidual} and densely connected convolutional networks~\cite{denseNet}. ICNNs employ such connections in order to restrict the energy function to be a convex function with respect to some parameters of the energy function. However, convexity gives a strong restriction on the expressivity of the energy function. In our experiments, we show that convexity constraints of ICNNs hinder the networks from generating visually better image inpainting results, and thus our method performs better than ICNNs.

\section{Proposed Method}
\label{sec:method}
\begin{figure*}
\begin{center}
\includegraphics[width=1.0\linewidth]{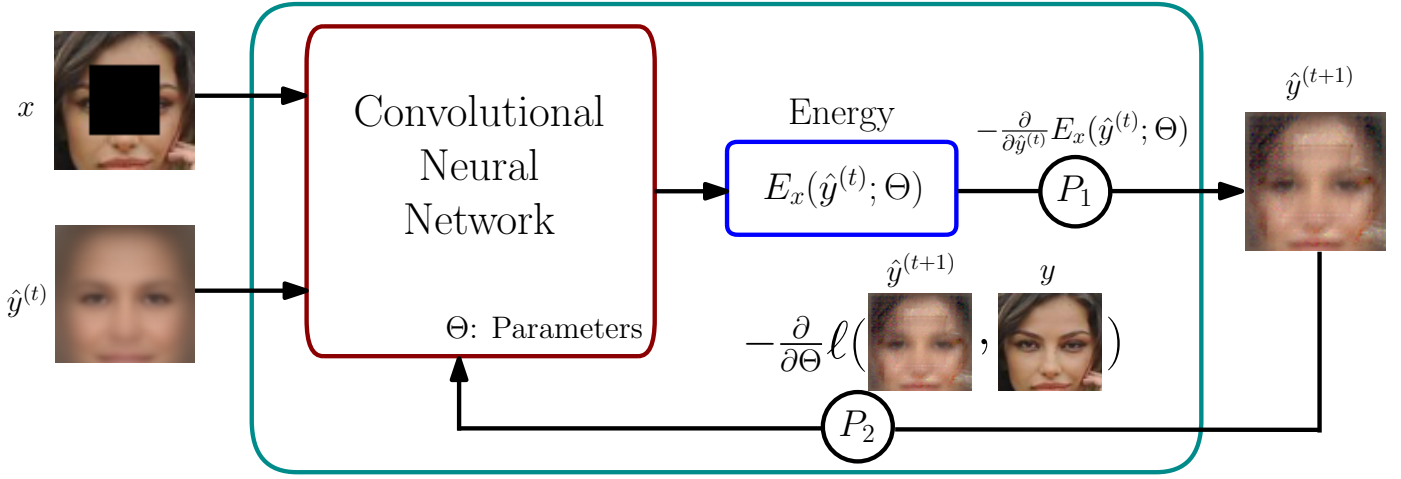}
\end{center}
   \caption{An overview of the proposed method. The network takes an occluded image $ x $ and an estimate $ \hat{y}^{(t)} $ of the inpainted, true image as inputs. 
   The operation $P_1$ performs gradient descent to compute the minimizer $\hat{y}$ to $E_x(\hat{y};\Theta)$ for a given $\Theta$. The number of iterations is set to 1 for simplicity of illustration. The operation $P_2$ updates the network parameter $\Theta$ by gradient descent to minimize 
   the loss function $ \ell(\hat{y}^{(t+1)}, y) $. For learning the network parameter $\Theta$, $P_1$ and $P_2$ are alternately performed, whereas for estimating the inpainted image, only $P_1$ is performed using learned $\Theta$.  }
\label{fig:frameworkOverview}
\end{figure*}

The proposed framework is illustrated in Fig.~\ref{fig:frameworkOverview}. The design of the network representing an energy function, learning procedures, and testing phase in our proposed framework are described in Section~\ref{sec:computationEnergy}, ~\ref{sec:learning}, and~\ref{sec:testing}, respectively.

\subsection{CNN Representing an Energy Function}
\label{sec:computationEnergy}
\begin{figure}[t]
\begin{center}
\includegraphics[width=1.0\linewidth]{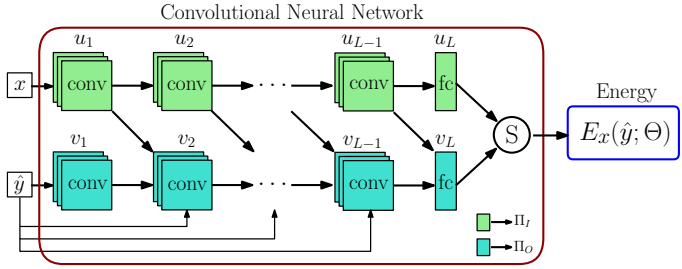}
\end{center}
  \caption{The CNN representing the energy function. It computes a scalar energy $E_{x}(\hat{y};\Theta)$, given an input image $x$ with missing regions and an estimate $\hat{y}$ of the true inpainted image. }
\label{fig:learningFramework}
\end{figure}

We use a CNN to represent an energy function $E_{x}(\hat{y}; \Theta)$, where $x$ is an input image with missing regions,  $\hat{y}$ is an estimate of the true image $y$ (i.e., the inpainted version of $x$), and $\Theta$ is the parameters (weights) of the CNN. Figure \ref{fig:learningFramework} shows its architecture. 
It contains two paths called input path $ \Pi_I $ and output path $ \Pi_O $, each of which has $ L $-layers. The image $ x $ is fed to $ \Pi_I $, and the image $ \hat{y} $ is fed to $ \Pi_O $. $ \Pi_I $ has parameters $ {\theta_{\Pi_I} = \{ W_{l,k}^{(u)}, b_{l,k}^{(u)}: k = 1,2,\ldots,K \}_{l=1}^L }$ and $ \Pi_O $ has parameters $ {\theta_{\Pi_O} = \{ W_{l,k}^{(v)}, W_{l,k}^{(u)}, W_{l,k}^{(z)}, b_{l,k}^{(v)}: k = 1,2,\ldots,K \}_{l=1}^L }$.
They comprise all the trainable parameters of the CNN, i.e., $ \Theta = \{ \theta_{\Pi_I}, \theta_{\Pi_O} \}$.

\medskip
\noindent
{\bf Input path $ \Pi_I $:} 
We compute the $ k^{th} $ feature map at the ${(l+1)^{st}}$ layer of the input path $ \Pi_I $ by
\begin{equation}
u_{l+1}^{k} = \sigma(W_{l,k}^{(u)} \ast u_{l}^{k} + b_{l,k}^{(u)}), 
\end{equation}
where $\ast$ denotes the convolution operation; $\sigma(\cdot)$ is an activation function; $ W_{l,k}^{(u)} $ and $ b_{l,k}^{(u)}$ denotes the weight matrix and the bias vector applied to the $ k^{th} $ feature map $ u_{l}^{k}$, ($k = 1, 2, \ldots, K$ and  $l = 1, 2, \ldots, L$). At the input layer $(l = 0)$, we set $u_{0}^{k} = x$. Following the convolutional layers, fully connected layers are used in $\Pi_I$. In Fig.~\ref{fig:learningFramework}, $u_l = [u_{l}^{1}, u_{l}^{2}, \dots, u_{l}^{K} ]$ denotes a tensor of the feature maps obtained at the $ l^{th} $ layer of $\Pi_I$.

\medskip
\noindent
{\bf Output path $ \Pi_O $:}
As shown in Fig.~\ref{fig:learningFramework}, there are connections from $\Pi_I$ to $\Pi_O$. Specifically, each layer $l(\geq 2)$ of this path has input connections from the lower layer output of $\Pi_I$ (i.e., $u_{l-1}$) in addition to the lower layer output (i.e., $v_{l-1}$). Furthermore, each layer except the last fully connected (FC) layer has a connection from the input $\hat{y}$, which  is rescaled to fit to the size of its input feature map.
Thus, the $ k^{th} $ feature map $ v_{l+1}^{k} $ is computed by
\begin{equation}
v_{l+1}^{k} = \sigma(W_{l,k}^{(v)} \ast v_{l}^{k} + W_{l,k}^{(u)} \ast u_{l}^{k} + W_{l,k}^{(z)} \ast z_{l}^{k} + b_{l,k}^{(v)}),
\end{equation}
where $z_{l}^{k}$ is the rescaled version of $ \hat{y} $ at the $l^{th}$ layer;
$W_{l,k}^{(v)}$ and $W_{l,k}^{(z)}$ are the weights applied on the feature maps $v_l^k$ and $z_l^k$, respectively  ($ k = 1, 2, \dots, K$ and $l = 1, 2, \ldots, L $). At the input layer $(l = 0)$ of the output path $ \Pi_O $, we set  $ W_{0,k}^{(z)} = 0 $, $ W_{0,k}^{(u)} = 0 $, and $ v_{0}^{k} = \hat{y} $. 

In order to compute the scalar value of the energy function $ E_{x}(\hat{y}; \Theta) $ for the inputs $ x $ and $ \hat{y} $ of the CNN, an operation (denoted by S) is employed following the top FC layers of the input path $ \Pi_I $ and the output path $ \Pi_O $ (see Fig.~\ref{fig:learningFramework}). The S-operation performs element-wise addition on the results obtained from the FC layers. Then, the output of the S-operation is reshaped to a scalar value of the energy.

\subsection{Learning the Energy Function}
\label{sec:learning}

\begin{algorithm}[t]
\SetAlgoLined
\KwIn{$ {\{(x_i, y_{i})\}}_{i=1}^{N}: $ Training set of image pairs.}
\KwIn{$ \Theta: $ Parameters of the network.}
$ \hat{y}^{(0)} \leftarrow \frac{1}{N}\sum_{i=1}^{N}y_i$\;
Initialize the parameters $ \Theta^{(0)} $\;
$ E^{(0)}_x \leftarrow E_{x}(\hat{y}^{(0)}; \Theta^{(0)}) $\;
\For{$ m \gets 0 $ \KwTo $ M $}{
Randomly choose an image pair $ (x, y) \in {\{(x_i, y_{i})\}}_{i=1}^{N} $ from the training set\;
\For{$ t \gets 0 $ \KwTo $ T $}{
$ \hat{y}^{(t+1)} \leftarrow \hat{y}^{(t)} - \alpha \frac{\partial}{\partial \hat{y}^{(t)}}E^{(t)}_x $\;
$ E^{(t+1)}_x \leftarrow E_{x}(\hat{y}^{(t+1)}; \Theta^{(m)}) $\;
}
$ \Theta^{(m+1)} \leftarrow \Theta^{(m)} - \lambda \frac{\partial}{\partial \Theta^{(m)}}\ell(\hat{y}^{(T)}, y) $\;
}
\caption{Our proposed training algorithm.
Learning rate used for optimization of energy $ E_{x}(\hat{y}; \Theta) $ in $T$ iterations is denoted by $ \alpha $, and learning rate used for optimization of parameters $ \Theta $ is denoted by $ \lambda $.}
\label{alg:spenipAlgorithm}
\end{algorithm}

We learn the energy function $E_x(\hat{y};\Theta)$ (i.e., its parameter $\Theta$) using $N$ pairs of an input image $x$ and its true image $y$. We denote the set of input images and that of the true images by $ {\mathcal{X} = {\{x_i\}}_{i=1}^{N}}$ and ${\mathcal{Y} = {\{y_{i}\}}_{i=1}^{N}}$, respectively. In order to determine $\Theta$, we consider the following constrained minimization problem:
\begin{equation}
\label{eq:constmin}
\min_\Theta \ell(\hat{y},y)\;\;\;\mbox{subject to}\;\;\;
\hat{y}=\argmin_{\hat{y}} E_x(\hat{y};\Theta).
\end{equation}
The underlying idea \cite{BelangerSPEN} is that we wish to obtain $\Theta$ such that the minimizer $\hat{y}$ to $E_x(\hat{y};\Theta)$ for a given $x$ and $\Theta$ provides the most similar image $\hat{y}$ to the ground truth $y$ for any pair of $x$ and $y$. 
We use the $\ell_1$ distance  between $ \hat{y} $ and  $ y $ for $l(\hat{y},y)$:
\begin{equation}
\label{eq:loss}
\ell(\hat{y}, y) = \|\hat{y} - y\|_1.
\end{equation}

The constrained minimization (\ref{eq:constmin}) can be converted to the following unconstrained minimization, if we can write the minimizer $\hat{y}$ to $E_x(\hat{y};\Theta)$ for a given $\Theta$ as $\hat{y}(\Theta)$: 
\begin{equation}
    \min_\Theta l(\hat{y}(\Theta),y).
\end{equation}
Thus, for a training sample ($x_i$,$y_i$), we first minimize $E_x(\hat{y};\Theta)$ for a fixed $\Theta$ with respect to $\hat{y}$ using gradient descent and then update $\Theta$ in the direction to minimize $l(\hat{y}(\Theta),y)$ using gradient descent. We iterate this pair of computation of $\hat{y}(\Theta)$ and update of $\Theta$ until convergence using training samples $\cal{X}$ and $\cal{Y}$. In Fig.~\ref{fig:frameworkOverview}, the computation of $\hat{y}(\Theta)$ is denoted by $P_1$ and the  updating of $\Theta$ is denoted by $P_2$.

The overall algorithm is given in Algorithm~\ref{alg:spenipAlgorithm}. For simplicity of explanation, a batch of size $ 1 $ is used here. The initial value $ \hat{y}^{(0)} $ of $\hat{y}$ is set to be the mean image of the ground truth images belonging to the set $\mathcal{Y}$. Then, the initial energy can be computed using an occluded image $ x $, the image $ \hat{y}^{(0)} $ and initial parameters $ \Theta^{(0)} $ of the network.
After the maximum number $ T $ of iterations of the update of $\hat{y}$ with a fixed $ \Theta=\Theta^{(m)} $ is achieved, the resulting $\hat{y}(\Theta)$ is used to update $ \Theta^{(m)} $ to minimize the loss function $ \ell(\hat{y}^{(T)}, y) $. The learning phase is completed after $M$ iterations.

\subsection{Inference for a Novel Image $x$}
\label{sec:testing}

Once the energy is learned, we can obtain an inpainted image for a novel input image $x$ with missing regions. This is done by minimizing the energy $E_x(\hat{y};\Theta)$ with respect to $\hat{y}$, while fixing $\Theta$ and $x$. In order to perform this minimization, we first initialize $\hat{y}$ with a mean image as in the first step of Algorithm 1. Then, we iteratively update $\hat{y}$ for $ t = 1, 2, \ldots, T $ according to line 7 of Algorithm 1. Our estimate is $\hat{y}^{(T)}$.

\section{Experimental Analyses}
\label{sec:results}

\begin{figure*}
\begin{center}
\includegraphics[width=1.\linewidth]{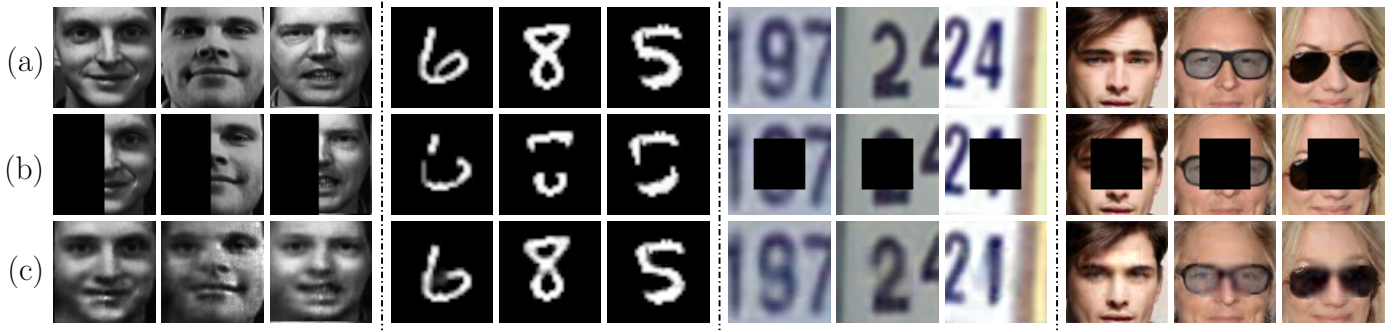}
\end{center}
   \caption{Examples of results obtained in image inpainting experiments using the test split of the Olivetti face, the MNIST, the SVHN, and the CelebA datasets, respectively. (a) Ground truth images, (b) occluded input images, and (c) our inpainting results.}
\label{fig:fullOcc}
\end{figure*}

\subsection{Implementation Details}
\label{sec:implementationDetails}

For implementation of the network in the proposed method, we used a simple CNN architecture. Our architecture consists of three convolutional and one fully connected layers in all our experiments. Rectified Linear Unit (ReLU) is employed as the activation function. We used learning rate $\alpha$ of 0.01 and momentum of 0.9 for energy update, and learning rate $\lambda$ of 0.001 for parameter update. Parameters were updated using the ADAM optimization algorithm~\cite{adam}. We implemented our proposed method using the TensorFlow framework. We used the same experimental settings for training and testing. Peak Signal to Noise Ratio (PSNR) metric was employed to quantitatively evaluate the inpainting results by
\begin{equation}
PSNR = 10 \cdot \log_{10} \frac{255^2}{\varepsilon},
\end{equation}
where $ \varepsilon $ is the mean squared error between the inpainted image and the ground truth image. PSNR results were averaged for samples belonging to the test set, and an average PSNR value was given for each experiment. Implementation details, hyperparameters and the code were made publicly available at \url{https://github.com/cvlab-tohoku/DSEBImageInpainting}.

\subsection{Datasets}
\label{sec:datasets}

We evaluate the proposed method using four datasets: The Olivetti face dataset~\cite{Olivetti}, the MNIST~\cite{mnist}, the Street View House Numbers (SVHN)~\cite{svhn}, and the Large-scale CelebFaces Attributes (CelebA)~\cite{celeba}. The Olivetti face dataset~\cite{Olivetti} contains 400 grayscale face images of size 64$ \times $64, and 50 images of the dataset were used for testing. The MNIST dataset~\cite{mnist} contains 70,000 grayscale images of size 28$ \times $28. We used 10,000 images of the MNIST dataset for testing, and all images were resized to 64$ \times $64. The SVHN dataset~\cite{svhn} consists of 99,289 RGB color images. We used 73,257 images for training and 26,032 images for testing. The images in the training and testing sets were resized to 64$ \times $64. The CelebA dataset~\cite{celeba} contains 202,599 RGB color face images. We used 2000 images for testing. All images belonging to the CelebA dataset were cropped to 64$ \times $64 at the center in order to extract regions that cover only faces. We performed the inpainting tests using two types of block occlusions:
\begin{itemize}
    \item Center block occlusion,
    \item Half block occlusion.
\end{itemize}
For center block occlusion tests, we created masks which cover approximately 25\% of the images. We located the masks at the center of images for the center block occlusion tests. Moreover, we performed half block occlusion tests on the Olivetti face dataset creating only left half occlusion masks in order to compare with the existing methods. Center block occlusion tests were performed using all datasets except the Olivetti face dataset. Half block occlusion tests were performed using only the Olivetti face dataset.

\subsection{Analyses using Grayscale Images}
We first provide test results using grayscale images which belong to the Olivetti face dataset in our analyses. The first three columns of Fig.~\ref{fig:fullOcc} depict the test results using the Olivetti face dataset. The results obtained using our proposed method for the half block occlusion inpainting task are shown in the last row of the figure. 
The results show that our proposed method generates images which are visually realistic and similar to ground truth images.

We exploited the same training and test splits, and minimized mean squared error (MMSE) in order to directly compare our results with the baseline methods~\cite{ICNN, SPN}. Table~\ref{table:olivettiMSEResults} shows the test MMSE results for our method and the state-of-the-art methods. As shown in the Table~\ref{table:olivettiMSEResults}, our proposed method outperforms the baseline methods using the Olivetti face dataset with a remarkable performance boost.

\begin{table}[h]
\caption{Mean squared errors computed for the half block occlusion inpainting task using the test split of the Olivetti face dataset.}
\begin{center}
\begin{tabular}{l c}
\hline
Method & Mean Squared Error \\
\hline
Sum Product Networks~\cite{SPN} & 942.0 \\
Input Convex Neural Networks (ICNNs) ~\cite{ICNN} & 833.0 \\
Our method & \textbf{497.35}\\
\hline
\end{tabular}
\end{center}
\label{table:olivettiMSEResults}
\end{table}

\begin{figure}[h]
\begin{center}
\includegraphics[width=0.9\linewidth]{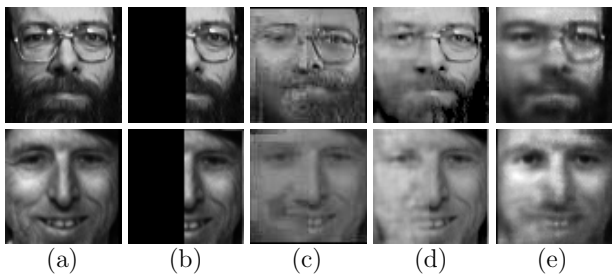}
\end{center}
   \caption{Comparison of test results for the half block occlusion inpainting task using the Olivetti face dataset. (a) Ground truth images, (b) occluded input images, (c) Sum Product Networks~\cite{SPN}, (d) ICNNs~\cite{ICNN}, (e) our method.}
\label{fig:olivettiComparisonResult}
\end{figure}

Fig.~\ref{fig:olivettiComparisonResult} shows a comparison of results obtained using the proposed method and the baseline methods \cite{ICNN, SPN} for the half block occlusion inpainting task on the Olivetti face dataset. The last column shows our results in the figure. The results show that our method generates images which are visually more similar to the ground truth images compared to the baseline methods. The second three columns of  Fig.~\ref{fig:fullOcc} depict our results obtained for the center block occlusion inpainting task using the test split of the MNIST dataset. The (c) column of the figure shows our inpainting results.

\subsection{Analyses using Color Images}

In this subsection, we compare our results with the state-of-the-art Context Encoders~\cite{pathakCVPR16context} and Semantic Image Inpainting~\cite{yeh2017semantic} methods. We used the same training and test splits, and masks that were used for analysis of the state-of-the-art methods \cite{pathakCVPR16context,yeh2017semantic} for a fair comparison.

In the third three columns of Fig.~\ref{fig:fullOcc}, we show examples of our results obtained for the center block occlusion inpainting task using the SVHN dataset. The last row of the figure shows our inpainting results. The results show that our method generates correct digits even if larger parts of digits are occluded by masks.

\begin{table}[h]
\caption{The results obtained for the center block occlusion inpainting task using the test split of the SVHN dataset.}
\begin{center}
\begin{tabular}{l c}
\hline
Method & PSNR (dB) \\
\hline
Context Encoders (CEs)~\cite{pathakCVPR16context} & 22.3 \\
Semantic Image Inpainting~\cite{yeh2017semantic} & 19.0 \\
Our method & \textbf{28.4}\\
\hline
\end{tabular}
\end{center}
\label{table:svhn}
\end{table}

\begin{figure}[h]
\begin{center}
\includegraphics[width=0.9\linewidth]{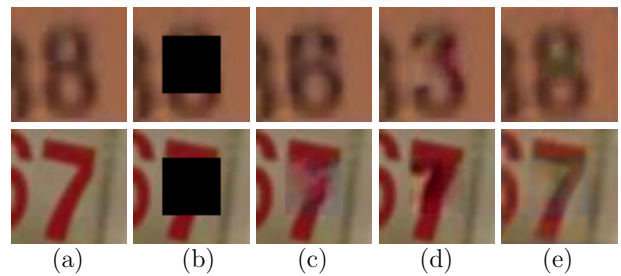}
\end{center}
   \caption{Comparison of test results for the center block occlusion inpainting task using the SVHN dataset. (a) Ground truth images, (b) occluded input images, (c) CEs~\cite{pathakCVPR16context}, (d) Semantic Image Inpainting~\cite{yeh2017semantic}, (e) our method.}
\label{fig:svhnResult}
\end{figure}

We compare PSNR results on the SVHN dataset for the center block occlusion inpainting task obtained using Context Encoders~\cite{pathakCVPR16context}, Semantic Image Inpainting~\cite{yeh2017semantic}, and our method in Table~\ref{table:svhn}. The ground truth images belonging to the test set of the SVHN dataset are used as reference to calculate the PSNR values. In the experiments, our method obtained the highest PSNR value providing  6.1 dB more than the PSNR provided by state-of-the-art method. The results also show that our method generates visually more similar images than the state-of-the-art methods.

Fig.~\ref{fig:svhnResult} shows samples of the results obtained for the center block occlusion inpainting task using the SVHN dataset. The  results indicate that our method generates visually more accurate and pleasing results compared to the state-of-the-art methods. The last three columns of Fig.~\ref{fig:fullOcc} show examples of results obtained for the center block occlusion inpainting task using the CelebA dataset. The last row of the figure shows our inpainting results. The results demonstrate that images generated by our method are similar to the ground truth images and visually realistic.

\begin{figure}[h]
\begin{center}
\includegraphics[width=0.9\linewidth]{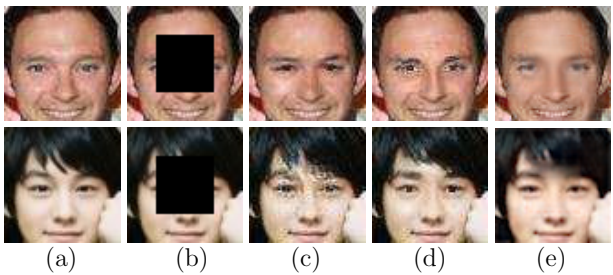}
\end{center}
   \caption{Comparison of test results for the center block occlusion inpainting task using the CelebA dataset. (a) Ground truth images, (b) occluded input images, (c) CEs~\cite{pathakCVPR16context}, (d) Semantic Image Inpainting~\cite{yeh2017semantic}, (e) our method.}
\label{fig:celebaResult}
\end{figure}

In Table~\ref{table:celeba}, we compare the PSNR results obtained using the CelebA dataset for the center block occlusion inpainting task. The ground truth images belonging to the testing set of the CelebA dataset were used as reference to calculate the PSNR values. These results indicate that the PSNR value of our method is 2.23 dB higher than that of the best of the state-of-the-art methods. Fig.~\ref{fig:celebaResult} shows a comparison of the results for the center block occlusion inpainting task using the CelebA dataset. The last column of the table given in the figure shows our inpainting results. The results show that our method generates inpainted images which are visually more similar to the ground truth images.

\begin{table}[h]
\caption{The results obtained for the center block occlusion inpainting task using the test split of the CelebA dataset.}
\begin{center}
\begin{tabular}{l c}
\hline
Method & PSNR (dB) \\
\hline
Context Encoders (CEs)~\cite{pathakCVPR16context} & 21.3 \\
Semantic Image Inpainting~\cite{yeh2017semantic} & 19.4 \\
Our method & \textbf{23.53}\\
\hline
\end{tabular}
\end{center}
\label{table:celeba}
\end{table}

\section{Conclusion}
In this work, we proposed a deep structured energy-based image inpainting method. The method is based on an energy based model which employs the energy function defined by a CNN. The proposed method can successfully inpaint the occluded region in images with visually more clear and realistic content compared to the baseline methods. Qualitative and quantitative results show that our proposed method achieves the state-of-the-art inpainting results by learning structural relationship between the patterns observed in images and occluded region of the images. In future work, we plan to employ the proposed method for other computer vision and pattern recognition tasks such as image denoising, super-resolution, image classification and object detection.

\bibliographystyle{IEEEtran}
\bibliography{bare_conf}

\end{document}